\documentclass[10pt,twocolumn,letterpaper]{article}

\usepackage{cvpr}
\usepackage{times}
\usepackage{epsfig}
\usepackage{graphicx}
\usepackage{amsmath}
\usepackage{amssymb}

\usepackage{multirow, dcolumn}
\usepackage{subcaption}
\usepackage{lipsum}
\usepackage{footnote}
\usepackage{siunitx}
\usepackage{makecell}
\usepackage{epstopdf}
\usepackage{fancyhdr}
\usepackage{xcolor}
\usepackage{gensymb}
\usepackage[normalem]{ulem}
\usepackage{soul}
\usepackage{verbatim}
\usepackage{booktabs}
\usepackage[pagebackref=true,breaklinks=true,letterpaper=true,colorlinks,bookmarks=false]{hyperref}

\cvprfinalcopy 

\def\cvprPaperID{23} 

\ifcvprfinal\fi

\begin{document}

\title{A Cyclically-Trained Adversarial Network for\\ Invariant Representation Learning}

\author{Jiawei Chen\\
Boston University\\
{\tt\small garychen@bu.edu}
\and
Janusz Konrad\\
Boston University\\
{\tt\small jkonrad@bu.edu}
\and
Prakash Ishwar\\
Boston University\\
{\tt\small pi@bu.edu}
}

\maketitle
\begin{abstract}
    Recent studies show that deep neural networks are vulnerable to adversarial examples which can be generated via certain types of transformations. Being robust to a desired family of adversarial attacks is then equivalent to being invariant to a family of transformations. Learning invariant representations then naturally emerges as an important goal to achieve which we explore in this paper within specific application contexts.
	Specifically, we propose a cyclically-trained adversarial network to learn a mapping from image space to latent representation space and back such that the latent representation is invariant to a {\it specified} factor of variation (e.g., identity). The learned mapping assures that the synthesized image is not only realistic, but has the same values for {\it unspecified} factors (e.g., pose and illumination) as the original image and a desired value of the specified factor. 
	Unlike disentangled representation learning, which requires two latent spaces, one for specified and another for unspecified factors, invariant representation learning needs only one such space.
	We encourage invariance to a specified factor by applying adversarial training using a variational autoencoder in the image space as opposed to the latent space. 
	We strengthen this invariance by introducing a cyclic training process (forward and backward cycle).
	We also propose a new method to evaluate 
	conditional generative networks. It compares how well different factors of variation can be predicted from the synthesized, as opposed to real, images.
	In quantitative terms, our approach attains state-of-the-art performance in experiments 
	spanning three datasets with factors such as identity, pose, illumination or style. 
	Our method produces sharp, high-quality synthetic images with little visible artefacts compared to previous approaches.
%
	%
	%
	
\end{abstract}

\vspace{-4ex}
\section{Introduction}
%
The performance of machine learning algorithms is usually related to the quality of internal data representation (features).
%
Thus, representation learning has been extensively studied in the fields of machine learning and artificial intelligence (AI).
\begin{figure}[!htb]
	\begin{subfigure}[!ht]{0.5\textwidth}
		\centering
		\includegraphics[width=0.95\linewidth]{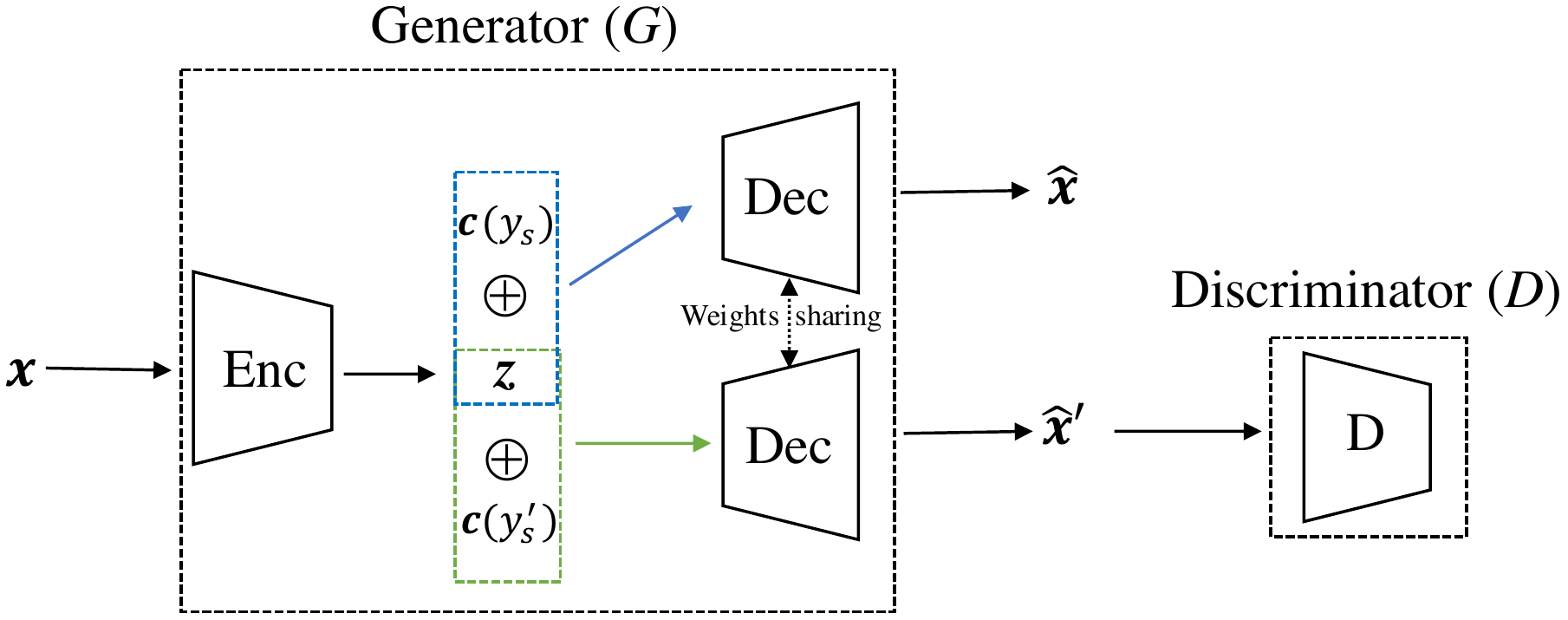}
		\caption{Forward cycle}
		\label{fig:fw_pass_diag} 
	\end{subfigure}%
	\quad
	\begin{subfigure}[!ht]{0.5\textwidth}
		\centering
		\includegraphics[width=0.8\linewidth]{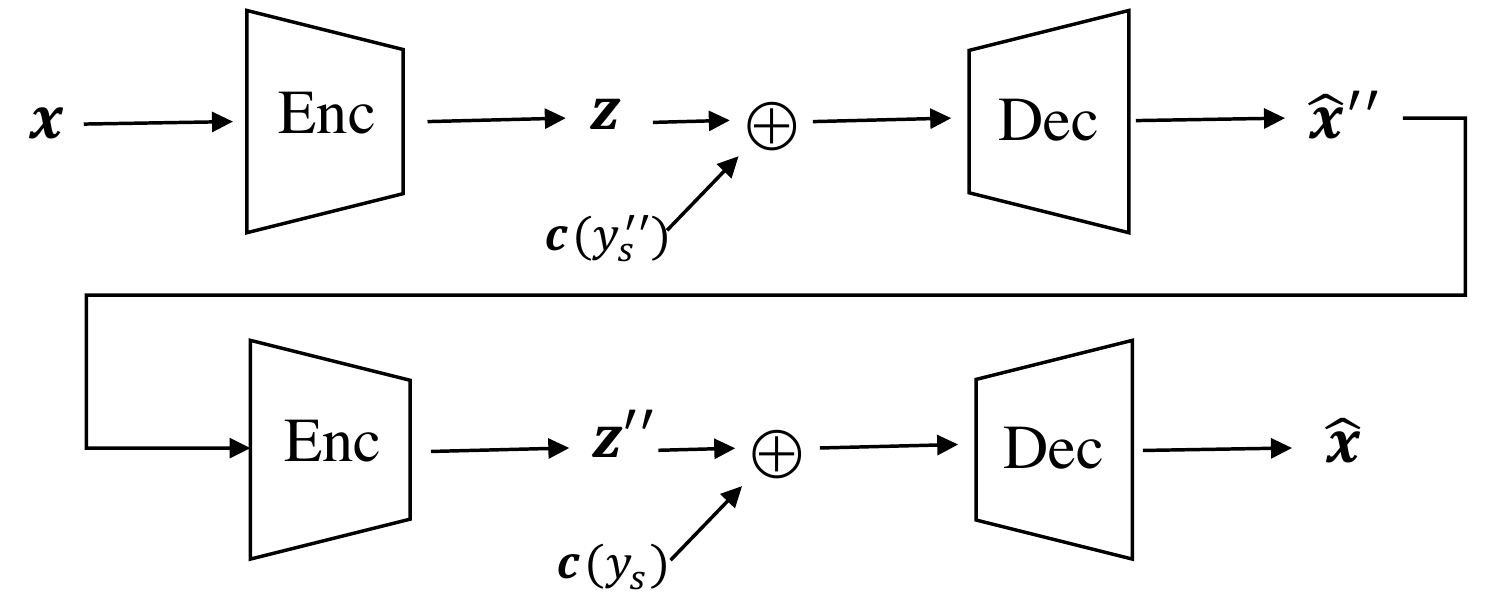}
		\caption{Backward cycle}
		\label{fig:bk_pass_diag}
	\end{subfigure}
	\vglue -0.2cm
	\caption{Diagram of the proposed model ($\oplus$ represents concatenation): (a) forward cycle in which training alternates between optimizing $G$ and $D$; (b) backward cycle that only optimizes $G$. Note: the label for image synthesis is denoted by $y'_s$ in the forward cycle and $y''_s$ in the backward cycle.
	}
	\label{fig:system_diag}
	\vglue -0.5cm
\end{figure}
Among the criteria for a ``good'' data representation, invariance to one or more specified factors of variation while simultaneously preserving other factors is an important property since it could benefit the end-task (e.g., classification) by factoring out irrelevant information~\cite{bengio2013}.
%
%
Immunity against imperceptible perturbations, which could mislead trained classifiers~\cite{goodfellow2014explaining,athalye2017synthesizing,liu2016delving}, can be attained by training a network to generate a representation that is invariant to such transformations~\cite{hwang2019puvae}.
Invariance to factors such as gender or race is also useful in applications where it is desirable that decisions not be biased towards or against a particular group. In such cases, the data representation must not contain group-identifying information, but must preserve other attributes.

Invariant representation learning has been studied in different domains, such as fair decision making~\cite{edwards2015censoring,calders2010three,zafar2015fairness}, privacy-preserving visual analytics~\cite{chen2016estimating,chen2017semi,chen2018vgan}, 
and domain adaptation~\cite{li2018domain,hoffman2013efficient,ganin2016domain}.
It is also closely related to disentangled representation learning, as both attempt to separate factors of variation in the data.
The major difference between them is that invariant representations eliminate unwanted factors in order to reduce sensitivity in the direction of invariance, while disentangled representations try to preserve as much information about the data as possible~\cite{bengio2013}.
In this work, we are interested in the problem of learning image representations that are invariant to certain \textit{specified} factors of variation, e.g., identity, while preserving other \textit{unspecified} factors, e.g., pose, expression, etc., using adversarial training. 
%
Our approach also enables us to explicitly set the value of the specified factor within a synthesized image.  

Generative models that are driven by an interpretable latent space, i.e., one where the data representations can be used to control certain characteristics of the outputs (e.g., create images from a particular class), are often preferable. 
Such models are useful in a variety of applications, e.g, automatic image editing~\cite{chen2018vgan,jin2017towards,lample2017fader}. 
Previous studies~\cite{chen2018vgan,lample2017fader,tran2017disentangled} combined the generative power of the Variational Auto-Encoder (VAE)~\cite{kingma2013auto} and the Generative Adversarial Network (GAN)~\cite{goodfellow2014generative} to learn an invariant latent space 
which enables
controlling a specified factor of variation in the synthesized samples. 
However, they either require training labels for both specified and unspecified factors of interest~\cite{chen2018vgan,tran2017disentangled} or are limited to binary attributes~\cite{lample2017fader}.

Our proposed framework also builds upon VAEs and GANs.
We combine the two by structuring the generator ($G$) in a conventional GAN as an encoder-decoder pair (see Fig.~\ref{fig:fw_pass_diag}).
In order to improve the invariance of data representations and quality of synthesized images, we introduce a forward-backward cyclic training process. 
During a forward cycle, the generator is given an input image $\mathbf{x}$ with a specified factor label  $y_s$. 
The encoder maps $\mathbf{x}$ to a latent representation $\mathbf{z}$, and the decoder is trained to reconstruct $\mathbf{x}$  based on $(\mathbf{z}, y_s)$ as well as synthesize a realistic image based on $(\mathbf{z}, y'_s)$, that can fool the discriminator into classifying it to class $y'_s$, where $y'_s$ is a generated class code. 
This encourages the encoder to reduce information about the specified factor in the latent representation as the specified factor of a generated sample is determined by the class code. 
Meanwhile, the encoder is also encouraged to pass information about the unspecified factors to the latent space to allow an accurate reconstruction.
However, a forward cycle alone does not prevent a \textit{degenerate} solution wherein information about the specified factor still leaks into the latent space,  but the decoder learns to ignore that information.
Therefore, in the backward cycle, we impose a further constraint in the latent space by explicitly minimizing the distance between two latent representations 
one for a real training image $\mathbf{x}$ associated with label $y_s$,  and another for a synthesized sample generated based on ($\mathbf{x}$, $y''_s$), where $y_s \neq y''_s$.
This forces the encoder to only encode unspecified factors within the latent space. 
The generator is additionally trained to reconstruct the real image from its synthetic version with the appropriate class code, which could benefit the image synthesis task. 

Once trained, our model becomes a conditional image generator that can synthesize novel images with the ability to change the specified factor value by tuning the class code (using the forward cycle only). 
%
In order to measure the quality of our model, we follow previous studies~\cite{hadad2018two,harsh2018disentangling} and conduct a subjective visual evaluation.  
We also propose a {\it quantitative} method to evaluate conditional generative models by measuring how well different factors of variation can be predicted in the synthesized images {\it via} pretrained attribute estimators. 

This paper makes the following contributions:
\begin{enumerate}
	%
	\item We develop a \textit{conditional Variational Generative Adversarial Network} for learning a representation that is invariant to a specified factor, while preserving other unspecified factors of variation.
	%
	\item We empirically verify the effectiveness of the proposed model in learning invariant representations \textit{via} a forward-backward cyclic training process on a number of datasets and tasks.
	%
	\item We propose a new quantitative method to evaluate the quality of conditional generative models and show that our model consistently produces better quality images compared to two state-of-the-art methods. 
\end{enumerate} 
\section{Related Work}
\noindent \textbf{Invariant Representation Learning:}
%
It has been extensively studied in various contexts and the related literature is vast. 
For instance, transformation-invariant feature learning has deep roots in computer vision; features are often designed for a specific case, e.g.,  rotation or scale invariance. 
Early examples include hand-crafted features such as HOG~\cite{dalal2005histograms} and SIFT~\cite{lowe1999object}.
More recently, deep Convolutional Neural Networks (CNNs) have been very effective in learning transformation-invariant representations
~\cite{cheng2016learning,cohen2016group,soatto2014visual}

Another line of research aims at building fair, bias-free classifiers that also attempt to learn representations invariant to ``nuisance variables'', which could potentially induce bias or unfairness~\cite{li2014learning,zafar2015fairness,edwards2015censoring,xie2017controllable}.  
%
One study proposed to obtain fairness by imposing $l_1$ regularization between representation distributions for data with different nuisance factors of variation~\cite{li2014learning}. 
The Variational Fair Auto-Encoder~\cite{zafar2015fairness} tackles the same task using a VAE with maximum mean discrepancy regularization.
Particularly relevant to our work are the methods proposed in~\cite{edwards2015censoring,xie2017controllable}, that also incorporate adversarial training in an auto-encoder framework. 
Our method differs in that we apply adversarial training in the image space instead of the latent space. 
%
This creates  better quality images.
%
%
Additionally, we improve the quality of invariance through cyclic training. 
%

Our work is also related to two recent studies~\cite{chen2018vgan,tran2017disentangled}. 
Both studies develop a fully-supervised method with adversarial training in the image space to learn invariant representations by factoring out nuisance variables for a specific task, e.g., identity-invariant expression recognition. 
However, both methods can only retain certain factors of variation (with labels available for training) in the representations. 
Another drawback is that in those models a discriminator is trained for each of the factors of variation so the number of model parameters grows linearly with the number of factors. 
In contrast, our model uses a single discriminator which only requires labels of the specified factor of variation.
Moreover, our model is designed to automatically capture all unspecified factors of the data into the representation with no need for corresponding labels.

\noindent \textbf{Disentangled Representation Learning:}
Invariant representation learning has a natural connection to disentangled representation learning, where the goal is to factorize different influencing factors of the data into different parts of its representation. 
In an early study, a bilinear model was proposed to separate content and style for face and text images~\cite{tenenbaum2000separating}. 
Another method used an E-M algorithm to discover the independent factors of variation of the underlying data distribution~\cite{ghahramani1995factorial}.
Later, unsupervised approaches to learn disentangled image representations were proposed~\cite{chen2016infogan,hu2018disentangling}. A purely-generative model was developed in \cite{chen2016infogan} but, unlike our model, it has no capacity to create an invariant representation for a given image.
A method proposed in \cite{hu2018disentangling} can learn an image representation that consists of a pre-defined number of disentangled factors of variation, but it has no control over which factors to learn.
%
%
%
Recent methods~\cite{mathieu2016disentangling,hadad2018two} proposed to combine auto-encoder with adversarial training to disentangle specified and unspecified factors of variation and map them onto separate latent spaces.
%
%
Indeed, the resulting unspecified representation is equivalent to an invariant representation that is disentangled from the specified factor. 
However, methods with sole pixel-wise reconstruction objective in the image space tend to produce blurry images.
Another recent work~\cite{harsh2018disentangling} proposed a cycle-consistent VAE  to disentangle the latent space into two complementary subspaces by using weak supervision (pairwise similarity labels).
%
%
It is related to our work in the sense that both methods constrain the latent space by adding a pair-wise distance between two latent representations (which are supposed to be close) into the cost function.  
The difference is that our method leverages adversarial training based on a single source of supervision, enabling
training with a single image in each iteration instead of a pair of images.
Moreover, we impose cycle-consistency loss in the image space as opposed to the latent space in~\cite{harsh2018disentangling}. Such modification, along with the additional adversarial loss in the image space, promotes generation of better quality invariant representations and  images.
%
%
%

It is also worthwhile to mention that our work is related to several previous works on image generation~\cite{larsen2015autoencoding,bao2017cvae} in terms of using auto-encoder and GAN. However, our goals are very different. 
The previous works mainly focus on developing image generation models, whereas our model is explicitly optimized to create invariant image representations. 
Once trained, our model becomes a conditional image generator. 
\section{Methodology}
Let $\mathbf{X}$ denote the image domain and $\mathbf{Y} = \{y_1,...,y_K\}$ 
%
a set of $K$ possible factors of variation associated with data samples in $\mathbf{X}$.
%
Given an image $\mathbf{x}\in\mathbf{X}$ and one specified factor $y_s$, where $y_s \in \{1,...,N_s \}$ and $N_s$ is the number of possible classes, our proposed approach has two objectives: 
1) to learn a latent representation $\mathbf{z}$ which is invariant to the specified factor but preserves the other unspecified factors of variation, and 
2) to synthesize a realistic sample
$\hat{\mathbf{x}}'$which has the same unspecified factors as $\mathbf{x}$ and a desired specified factor value which is determined by an input class code $\mathbf{c}(y'_{s})$, where $y'_{s} \in \{1,...,N_s \}$ is generated from a distribution $p(y'_{s})$ and $\mathbf{c}(\cdot)$ is a one-hot encoding function.
For simplicity, we consider here the case where $y_s$ is categorical, but our approach can be extended to continuous $y_s$.

\noindent \textbf{Generator:}
We structure the generator in the proposed model as an encoder-decoder pair (Fig.~\ref{fig:system_diag}).
The encoder ($Enc$) aims to create a low-dimensional data representation $\mathbf{z} = Enc(\mathbf{x})$ {\it via} a randomized mapping $\mathbf{z} \sim p(\mathbf{z}|\mathbf{x})$ parameterized by the weights of the encoder's neural network $\theta_{enc}$. 
On the other hand, the decoder ($Dec$) is a neural network with weights $\theta_{dec}$.
It is responsible for learning a mapping function $\hat{\mathbf{x}}' \sim p(\mathbf{x}|\mathbf{z}, \mathbf{c}(y'_s))$ that can map the latent representation $\mathbf{z}$ in combination with class code $\mathbf{c}(y'_s)$ back to the image space.  
The latent space is regularized by imposing a prior distribution, in our experiments a normal distribution $r(\mathbf{z}) \sim \mathcal{N}(\mathbf{0},\mathbf{I})$.

\noindent \textbf{Discriminator:}
Different from the discriminators in conventional GANs, the discriminator $D$ in our model is a multi-class classifier represented by a neural network with weights $\theta_{dis}$.
The outputs of the discriminator $D(\mathbf{x})\in \mathbb{R}^{N_s+1}$ are the predicted probabilities of each class corresponding to $N_s$ different values of the specified factor and an additional ``fake'' class for synthesized images. 

\noindent \textbf{Forward cycle:}
%
%
First, we sample an image $\mathbf{x}$ from the training set and pass it through the encoder to generate a latent representation $\mathbf{z}$. 
The decoder is trained to produce a reconstruction of the input $\hat{\mathbf{x}} \sim p(\mathbf{x}|\mathbf{z}, \mathbf{c}(y_s))$ and also to synthesize a new data sample $\hat{\mathbf{x}}' \sim p(\mathbf{x}|\mathbf{z}, \mathbf{c}(y'_s))$ that can fool the discriminator $D$ into classifying it as the specified class $y'_s$.
Specifically, the weights of the generator network are adjusted to {\it minimize} the following cost function:
%
%
\begin{align}
\mathcal{L}^{fw}_{G}&(G,D) = \nonumber \\
& -\lambda^{G}_1 E_{\mathbf{x} \sim p(\mathbf{x}), y'_s \sim p(y'_s)}\big[ \log D_{y'_s}\big(G(\mathbf{x}, \mathbf{c}(y'_s))\big) \big] + \nonumber \\
& \lambda^{G}_2 E_{(\mathbf{x},y_s) \sim p(\mathbf{x}, y_s)}\big[ ||\mathbf{x} - G(\mathbf{x}, \mathbf{c}(y_s))||_2^2 \big] 
+ \nonumber \\
& \lambda^{G}_3 KL(p(\mathbf{z}|\mathbf{x}) || r(\mathbf{z}))
\label{eq:gen_cost_fw}
\end{align}
where $p(\mathbf{x},y_s)$ denotes the joint distribution of the real image and the specified factor in the training data, $p(\mathbf{x})$ is the corresponding marginal distribution of the real image, $p(y'_s)$ is a distribution of the specified factor used to synthesize a ``fake'' image,
%
%
$D_i$ is the predicted probability of the $i$-th class, and $\lambda^{G}_1, \lambda^{G}_2, \lambda^{G}_3$ are weighting factors. 
%
%

The discriminator aims to correctly classify a real training sample $\mathbf{x}$ to its ground-truth class value $y_s$ of the specified attribute but, when given a synthetic sample $\hat{\mathbf{x}}'$ from the generator, it attempts to classify it as fake. 
This is accomplished by adjusting the weights of the discriminator by {\it maximizing} the following cost function:
\begin{align}
\mathcal{L}^{fw}_{D}&(G,D) = \nonumber \\
& \lambda^{D}_1 E_{(\mathbf{x},y_s) \sim p(\mathbf{x}, y_s)}[ \log D_{y_s}(\mathbf{x})] + \nonumber \\
& \lambda^{D}_2 E_{\mathbf{x} \sim p(\mathbf{x}), y'_s \sim p(y'_s)}[ \log D_{N_s+1}(G(\mathbf{x}, \mathbf{c}(y'_s)))] 
\label{eq:disc_cost_fw}
\end{align}
where $\lambda^{D}_1$ and $\lambda^{D}_2$ are tuning parameters.

%
The weights of the networks in $G$ and $D$ are updated in an alternating order.
Over successive training steps, $G$ learns to fit the true data distribution and reconstruct the input image as well as synthesize realistic images that can fool $D$. 
The generator objective (second term in Eq.~(\ref{eq:gen_cost_fw}))
encourages the encoder to pass as much information about the unspecified factors as possible to the latent representation. 
Since the class code $\mathbf{c}$ determines the specified factor value of $\hat{\mathbf{x}}'$, the encoder is also encouraged to eliminate information about the specified factor of $\mathbf{x}$ in the latent representation. 
The encoder may, however, fail to disentangle the 
specified and unspecified factors of variation and the decoder may still learn to synthesize images according to the class code $\mathbf{c}$ by ignoring any \textit{residual} information about the specified factor that is contained within the representation.
%
%
To avoid such a \textit{degenerate} solution, we use a backward cycle to further constrain the latent space.  

\noindent \textbf{Backward cycle:}
This  cycle requires a synthesized image $\hat{\mathbf{x}}''$ of class $y''_s$ generated from a real image $\mathbf{x}$ of class $y_s$. 
We intentionally choose $y''_s \neq y_s$ so that $\hat{\mathbf{x}}''$ and $\mathbf{x}$ carry different specified factor values.
Two latent representations $\mathbf{z} = Enc(\mathbf{x})$ and $\mathbf{z}'' = Enc(\hat{\mathbf{x}}'')$ can be computed by passing, respectively,  $\mathbf{x}$ and $\hat{\mathbf{x}}''$ through the encoder.
If the encoder fails to transmit information about unspecified factors from the input to its latent representation, or if it retains considerable information about the specified factor in the latent space, then $\mathbf{z}$ and $\mathbf{z}''$ are expected to have a large pair-wise distance. In other words, if the latent space only maintains information about the unspecified factors,   $\mathbf{z}$ should be equivalent to $\mathbf{z}''$.
In addition, we would like
to encourage the generator to reconstruct the input $\mathbf{x}$ from its synthetic version $\hat{\mathbf{x}}''$ in combination with a class code $\mathbf{c}(y_s)$ that encodes the ground-truth label of the specified factor of $\mathbf{x}$.
These considerations motivate optimizing the generator in the backward cycle by minimizing the following cost function:
\begin{align}
\mathcal{L}^{bw}_{G} &=  E_{(\mathbf{x}, y_s) \sim p(\mathbf{x}, y_s), y''_s \sim p(y''_s)} \big[ \lambda^{bw}_1||\mathbf{z} - \mathbf{z}''||_1  + \nonumber \\
& \lambda^{bw}_2 ||\mathbf{x} - G(\hat{\mathbf{x}}'', \mathbf{c}(y_s))||_2^2\big]
\label{eq:gen_cost_bw}
\end{align}
%
where $ \lambda^{bw}_1$ and $ \lambda^{bw}_2$ are two weighting factors. The first term in Eq.~(\ref{eq:gen_cost_bw}) penalizes the generator if $\mathbf{z}$ is not close to $\mathbf{z}''$. 
The second term encourages the synthesized $\hat{\mathbf{x}}$ to resemble $\mathbf{x}$.

Essentially, the forward cycle translates $\mathbf{x}$ to a synthetic image $\hat{\mathbf{x}}'' = G(\mathbf{x}, \mathbf{c}(y''_s))$ followed by a backward transform $\hat{\mathbf{x}} = G(\hat{\mathbf{x}}'', \mathbf{c}(y_s))$, such that $\hat{\mathbf{x}} \simeq \mathbf{x}$.
This cyclic training process assists the model in generating good quality images and 
further encourages invariance to the specified factor in the latent space 

%
%
%
%
%
%
%

\section{Experimental Evaluation}
We evaluate the performance of the proposed model on three image datasets: 3D Chairs~\cite{aubry2014seeing}, YaleFace~\cite{KCLee05} and UPNA Synthetic~\cite{ariz2016novel}.
We first conduct a quantitative evaluation of the degree of invariance in the latent space by training dedicated neural networks (one per factor) to predict the values of the specified and certain unspecified factors (that have ground-truth labels) from the latent representation.
The factor prediction accuracies quantify how much information about each factor has been preserved in the latent representation. 
If the model succeeds in eliminating all information about the specified factor and preserving all information about unspecified factors, 
we should expect the prediction accuracy for the specified factor to be close to pure chance and the prediction accuracies for the unspecified factors to be nearly perfect. 
We also evaluate the quality of the image generation process. 
Unlike previous works~\cite{hadad2018two,harsh2018disentangling}, which only provide a qualitative evaluation through visual inspection of the synthesized images, we propose a new method to quantitatively assess the ability of a conditional generative model to synthesize realistic images while preserving unspecified factors.
In Section~\ref{sec:quality_img_gen}, we present details of the proposed evaluation method and associated experimental results.

We compare our model with two state-of-the-art methods~\cite{hadad2018two,harsh2018disentangling} that learn to produce, for a given input image, two latent vectors (as opposed to just one in our method). One of the latent vectors captures information related to the unspecified factors of variation and is, in an ideal scenario, devoid of any information related to the specified factor of variation. This latent vector is the counterpart of the latent invariant representation in our method.
For synthesizing an image with a desired value for the specified factor, the methods in \cite{hadad2018two,harsh2018disentangling} require an additional surrogate image which has the desired value for the specified factor. They would then \textit{substitute} the latent vector of the specified factor in the original image with that of the surrogate image and then decode the result. Our approach, in contrast, uses a class code (as opposed to a surrogate image) to explicitly set the value of the specified factor in the synthesized image. 
In our experiments, we compare the latent vectors for the unspecified factors from the competing methods and the latent representation from our method in terms of their ability to predict the specified and unspecified factors which indicates the quality of invariance.
We used the publicly-available source code to implement both benchmarks, but slightly modified their network architectures to ensure that all three competing models have similar numbers of parameters.
We also did parameter tuning for each method for each of the three datasets.

\subsection{Datasets}
\noindent \textbf{3D Chairs:} This dataset includes 1,393 3D chair styles rendered on a white background from 62 different viewpoints that are indexed by two values of angle $\theta$ and $31$ values of angle $\phi$. 
Each image is annotated with the chair identity 
indicating its style as well as viewpoint $(\theta,\phi)$.
For each chair style, we randomly picked 50 images (out of 62) to populate the training set, and used the remaining 12 images in the testing phase. This gives, in total,  69,650 images in the training set, and 16,716 images in the test set.

\noindent \textbf{YaleFace:} This dataset consists of gray-scale frontal face images of 38 subjects under 64 illumination conditions. 
In our experiments, we randomly chose 54 images (out of 64) from each subject for training, and used the rest as the test set for performance evaluation. 

\noindent \textbf{UPNA Synthetic:} 
This is a synthetic human head-pose database. 
It consists of 12 videos for each of 10 subjects; 120 videos in total with 38,800 frames. 
Ground-truth \textit{continuous} head pose angles (yaw, pitch, roll) are provided for each frame. 
We randomly selected $85\%$ of the frames from each video for each subject for the training and used the remaining $15\%$ for testing. 

For computational efficiency, in our experiments, we resized each RGB image to $64\times64$-pixel resolution for all three datasets.
Table~\ref{tbl:datasets_des} summarizes the specified and unspecified factors of variation that we investigate across the three datasets.  

\begin{table}[!thb]
\vglue -0.20cm
\caption{Specified and unspecified factor(s) of variation investigated in the three datasets. }
\vglue -0.30cm
\label{tbl:datasets_des}
\centering
\setlength\extrarowheight{1pt}
\resizebox{1\columnwidth}{!}{%
\begin{tabular}{ c c c }
\Xhline{1pt}
\multirow{1}{*}{\textbf{Dataset}}	
&   \multicolumn{1}{c}{\textbf{Specified factor}} 
&   \multicolumn{1}{c}{\textbf{Unspecified factor(s)}} \\
\hline		
3D Chairs & Chair style & View orientation ($\theta, \phi$) \\
\hline
YaleFace  & Identity &
Illumination Cond.\\
\hline
UPNA Synthetic & Identity & Head pose\\
\Xhline{1pt}
\end{tabular}}
\vglue -0.4cm
\end{table}
\begin{table*}[!htb]
\caption{Evaluation of the quality of invariance of representations generated by the competing models on 3D Chairs, YaleFace and UPNA Synthetic datasets. Classification performance is measured using CCR. Regression performance is measured using MAE and standard deviation.  $\uparrow$ means higher is better. $\downarrow$ means lower is better.  }
\vglue -0.1cm
\label{tbl:quality_invariance}
\centering
\resizebox{0.8\textwidth}{!}{
\setlength\extrarowheight{3pt}
\begin{tabular}{c c c c c c c }
	\toprule
	\multirow{2}{*}{\textbf{Datasets}} &  \multirow{2}{*}{\textbf{Factors of variation}} & \multicolumn{5}{ c }{\textbf{Methods}}  \\ 
	\cline{3-7}
	&   &  Random  guess/ & \multirow{2}{*}{~\cite{hadad2018two}}
	& \multirow{2}{*}{~\cite{harsh2018disentangling}} & \multirow{2}{*}{Ours }
	& \multirow{2}{*}{Ours w/o b.w. cycle} \\
	& & Median  & & & & \\
	\cline{1-7}
	\multirow{3}{*}{3D Chairs} & Chair Style $\downarrow$ & 0.07$\%$ & 0.77$\%$ & 0.70$\%$ & 0.79$\%$ & 3.21$\%$ \\
	& $\theta$  $\uparrow$ & 50$\%$ & 68.92$\%$& 64.22$\%$ & 78.17$\%$  & 74.37$\%$ \\
	& $\phi$ $\uparrow$ & 3.22$\%$ & 50.23$\%$ & 43.75$\%$ & 71.90$\%$ & 69.45$\%$ \\
	\cline{1-7}
	\multirow{2}{*}{YaleFace} &Identity $\downarrow$& 2.63$\%$ & 4.68$\%$ & 5.50$\%$ & 6.97$\%$ & 12.36$\%$ \\
	&  Illumination Cond. $\uparrow$ & 1.56$\%$ & 77.80$\%$ & 32.36$\%$ & 85.50$\%$ & 85.40$\%$ \\
	\cline{1-7}
	\multirow{4}{*}{UPNA Synthetic} & Identity $\downarrow$ & 10$\%$ & 15.80$\%$ & 18.83$\%$ & 18.05$\%$ & 33.40$\%$ \\
	& Yaw $\downarrow$ & 5.10$^{\circ}\pm$6.70$^{\circ}$ & 2.77$^{\circ}\pm$2.00$^{\circ}$ & 2.42$^{\circ}\pm$2.52$^{\circ}$  & 2.12$^{\circ}\pm$2.12$^{\circ}$ & 2.10$^{\circ}\pm$2.08$^{\circ}$  \\
	& Pitch $\downarrow$ & 4.98$^{\circ}\pm$5.02$^{\circ}$  & 2.43$^{\circ}\pm$2.10$^{\circ}$  & 2.88$^{\circ}\pm$2.71$^{\circ}$  & 2.23$^{\circ}\pm$2.10$^{\circ}$ & 2.20$^{\circ}\pm$2.06$^{\circ}$ \\
	& Roll $\downarrow$ & 4.68$^{\circ}\pm$6.88$^{\circ}$ & 1.19$^{\circ}\pm$1.43$^{\circ}$  & 1.65$^{\circ}\pm$2.35$^{\circ}$  & 1.16$^{\circ}\pm$1.24$^{\circ}$ & 1.29$^{\circ}\pm$1.43$^{\circ}$ \\
	\bottomrule
	
\end{tabular}}	
\vglue -0.5cm
\end{table*}

\subsection{Quality of invariance}
We follow previous methodology~\cite{harsh2018disentangling,hadad2018two} and train dedicated neural network estimators to predict the specified and unspecified factors of variation based on the learned latent representations generated by each competing model.
We use correct classification rate (CCR) and mean absolute error (MAE) to measure the performance of classification tasks and regression tasks, respectively.
Table~\ref{tbl:quality_invariance} summarizes the performance of each model on the three image datasets. 

In the 3D Chairs dataset, we regard chair style as the specified factor and the viewing orientation angles as the unspecified factors.
Since both orientation angles are discrete, we treat viewing orientation estimation as a classification problem. As shown in Table~\ref{tbl:quality_invariance}, all three competing models manage to reduce the style information contained within the latent representation to a large extent (very low style prediction CCR values). However the proposed model (with backward cycle) outperforms the benchmark models, in terms of the ability to predict the viewing orientation angles, by a large margin (about $11$--$28\%$ CCR improvement for $\phi$ and $9$--$13\%$ CCR improvement for $\theta$).
We also note that the backward cycle significantly improves invariance, e.g., style prediction CCR decreases from 3.21$\%$ to 0.79$\%$.

For the YaleFace dataset, subject identity is considered as the specified factor and illumination condition as the unspecified factor of variation. 
We first observe that the identification performance of the three models is comparable and close to a random guess, which suggests the competing models perform equally well in creating representations that are invariant to identity.
For the recognition of illumination condition, the classification CCR for our model is $85.50\%$, which again surpasses the two benchmark CCRs by about $8\%$ and $53\%$ in accuracy.
Such large performance gaps suggest that the invariant representation learned by our model is better, than the competing alternatives, in preserving information about unspecified factors of variation.  
Furthermore, we observe that the backward cycle helps reduce the identification CCR by about 5$\%$, thus confirming its usefulness.

In the case of UPNA Synthetic dataset, the specified and unspecified factors of variation used in evaluation are subject identity and head pose, respectively. 
%
Head pose is defined as a three-dimensional angular value (yaw, pitch, roll) in continuous space. 
Thus, we train neural-network based regressors to estimate head pose and report the mean and standard deviation of the absolute errors for yaw, pitch and roll angles separately. 
%
%
In terms of identification accuracy, the performance of the three methods is similar (no more than $3\%$ difference in CCR or about 2-3 times that of a random guess).
We also notice that the backward cycle greatly promotes invariance as it helps to reduce identification CCR from 33.40$\%$ to 18.05$\%$.
As for head-pose estimation, we use ``Median'' estimate as a baseline, i.e., the median value of ground truth across the entire training set. 
We note that our model slightly, but consistently, outperforms the benchmarks, and significantly outperforms the median estimate. 
This once again confirms the effectiveness of our model in preserving information pertaining to the unspecified factors in the latent representation while discarding information related to the specified factor.

\subsection{Quality of image generation}
\label{sec:quality_img_gen}
%

Many studies have proposed measures to evaluate generative models for image synthesis.
Some of them attempt to quantitatively evaluate models while some others emphasize qualitative approaches, such as user studies (e.g., visual examination).
However, such subjective assessment may be inconsistent and not robust as human operators may fail to distinguish subtle differences in color, texture, etc.
%
In addition, such a measure may favor models that can merely memorize training samples. 
%
In terms of quantitative methods, some studies proposed to use measures from image quality assessment literature such as SSIM, MSE and PSNR. 
However, they require a corresponding reference real image for each synthesized one. 
Other widely-adopted reference-free quantitative measures like Inception Score~\cite{salimans2016improved} and Fr\'echet Inception Distance~\cite{heusel2017gans} are designed for generic GANs. 
Thus, they are not suitable for conditional models that aim to generate samples from a particular class.
Several quantitative evaluation methods have been proposed for conditional generative models~\cite{choi2018stargan,zhang2016colorful}.
For example, ~\cite{zhang2016colorful} proposed to feed fake colorized images (of real grayscale images) to a classifier that was trained on real color images. 
If the classifier performs well, this indicates that the colorization is accurate. 
In contrast to these works, we evaluate multiple objectives simultaneously: one to evaluate invariance to a target attribute and others to evaluate the preservation of un-specified attributes.
%
%
%

Inspired by the previous studies that use an off-the-shelf classifier to assess the realism of synthesized data,
we propose a quantitative method that utilizes a number of attribute estimators to evaluate the quality of conditional generative models. 
The intuition is that a good generative model for learning an invariant/disentangled representation should have the capability to explicitly and accurately control the specified factor value when it generates a novel image.
Furthermore, it should precisely transfer the other unspecified factors of variation from the source image to its synthetic version.
Therefore, we can evaluate a model by measuring how well the different factors of variation in the synthesized images can be predicted {\it via} estimators that are pretrained on the real images. 

Specifically, we train a number of attribute estimators $\mathcal{F}^j$, where $j \in \{1,...,K\}$, on the original training sets of real images.
For each (real) test image $\mathbf{x}$ having specified and unspecified factors of variation $y_j$, $j\in \{1,\ldots,K\}$, we synthesize a new version $\hat{\mathbf{x}}' = G(\mathbf{x},\mathbf{c}(y'_s))$ using the generator, where $y'_s$ is sampled at random, independently of $\mathbf{x}, y_s$, from a distribution $p(y'_s)$. 
The image $\hat{\mathbf{x}}'$ thus synthesized is passed to the pretrained estimators to obtain a prediction for each attribute (whether specified or unspecified). 
If a factor of variation $y_j$ is categorical, then $\mathcal{F}^j(\hat{\mathbf{x}}')$ is a probability distribution over the set of all possible values that factor can take. In particular, $\mathcal{F}^j_{y_j}(\hat{\mathbf{x}}') = p(y_j|\hat{\mathbf{x}}')$.
If $y_j$ is continuous, then $\hat{y_j} := \mathcal{F}^j(\hat{\mathbf{x}}')$ is a numerical value which should be approximately equal to $y_j$.
In order to quantify performance, we introduce the following \textit{Generator Label Score} (\textit{GLS}) for both discrete and continuous factors of variation.  For a categorical unspecified factor $y_j$, 
\begin{align*}
GLS :=  
E_{(\mathbf{x},y_j)\sim p(\mathbf{x},y_j), y'_s \sim p(y'_s)}
\big[\mathcal{F}^j_{y_j} \big(G(\mathbf{x},\mathbf{c}(y'_s))\big)\big]
\end{align*}
whereas for a categorical specified factor $y_s$, 
\begin{align*}
GLS :=  
E_{\mathbf{x}\sim p(\mathbf{x}), y'_s \sim p(y'_s)}
\big[\mathcal{F}^j_{y'_s} \big(G(\mathbf{x},\mathbf{c}(y'_s))\big)\big]. 
\end{align*}
For a quantitative unspecified factor $y_j$,
\begin{align*}
GLS :=  
E_{(\mathbf{x},y_j)\sim p(\mathbf{x},y_j), y'_s \sim p(y'_s)}
||\mathcal{F}^j\big(G(\mathbf{x},\mathbf{c}(y'_s))\big) - y_j ||^p 
\end{align*}
whereas for a quantitative unspecified factor $y_s$,
\begin{align*}
GLS :=  
E_{\mathbf{x}\sim p(\mathbf{x}), y'_s \sim p(y'_s)}
||\mathcal{F}^j\big(G(\mathbf{x},\mathbf{c}(y'_s))\big) - y'_s ||^p.
\end{align*}
For a good conditional generative model, the value of \textit{GLS} should be high for every categorical factor of variation (specified or unspecified) and low for every quantitative factor.
If the relative importance of each attribute is known, \textit{GLS} values can be converted to a single value.
Although quantitative, $GLS$ need not correlate well with the subjective quality of synthesized images as perceived by humans.

In order to compute \textit{GLS}, we use the three competing models to create, separately, synthetic versions of test images for each dataset.  
For the proposed model, the input image $\mathbf{x}$ and class code $\mathbf{c}$ provide the necessary information about unspecified and specified factors, respectively. 
Thus, we synthesize a new version for each test image by passing it through the generator in combination with a randomly-generated class code.
For the benchmark methods, we follow the procedure described in the respective papers to generate new images.
In order to generate a new sample, we combine the unspecified latent representation of a test image and the specified latent representation of another image randomly picked from the same test set.

\begin{figure*}[!ht]
	\begin{subfigure}[b]{1\textwidth}
		\centering
		\includegraphics[width=.23\textwidth]{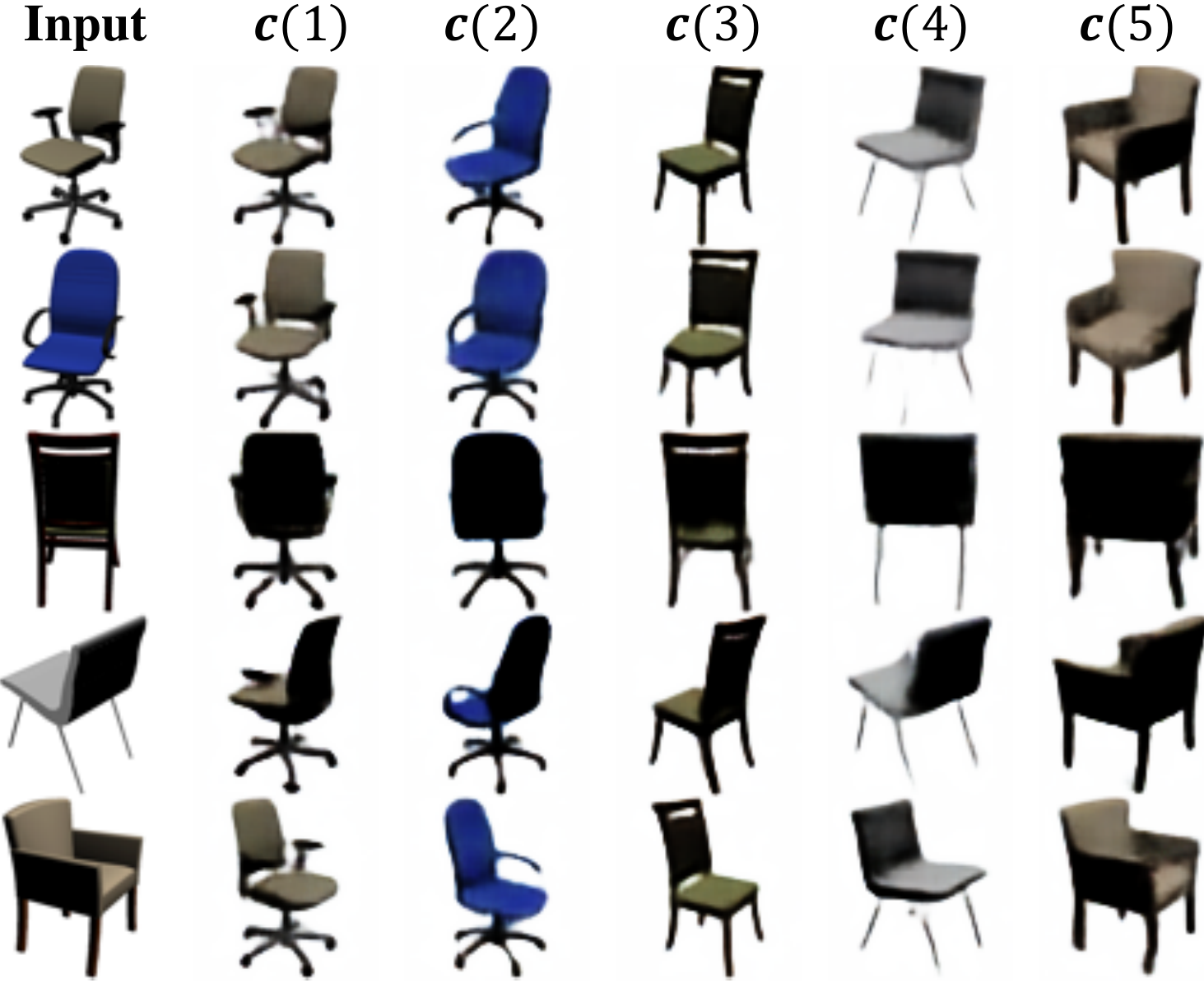}\qquad
		\includegraphics[width=.23\textwidth]{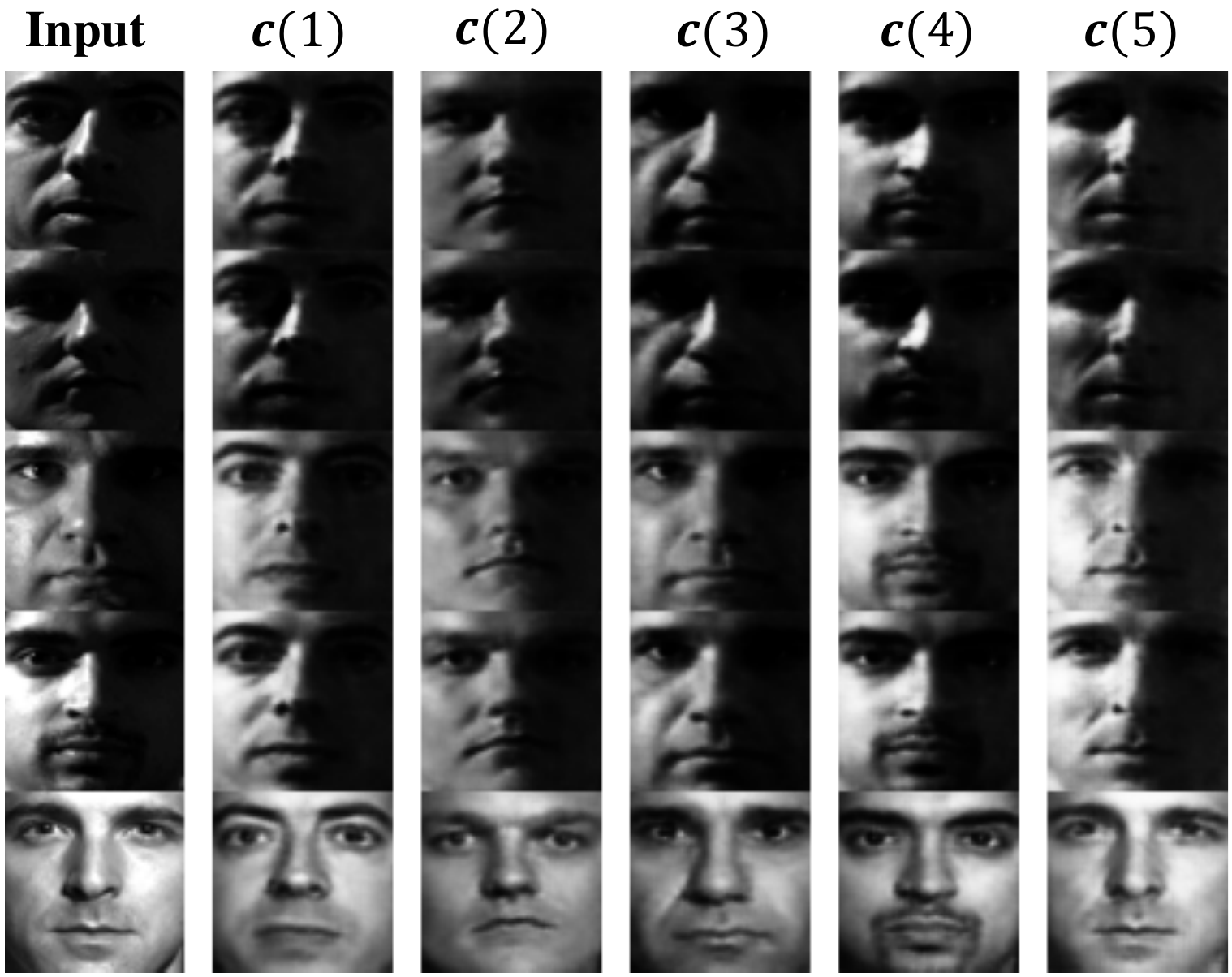}\qquad
		\includegraphics[width=.24\textwidth]{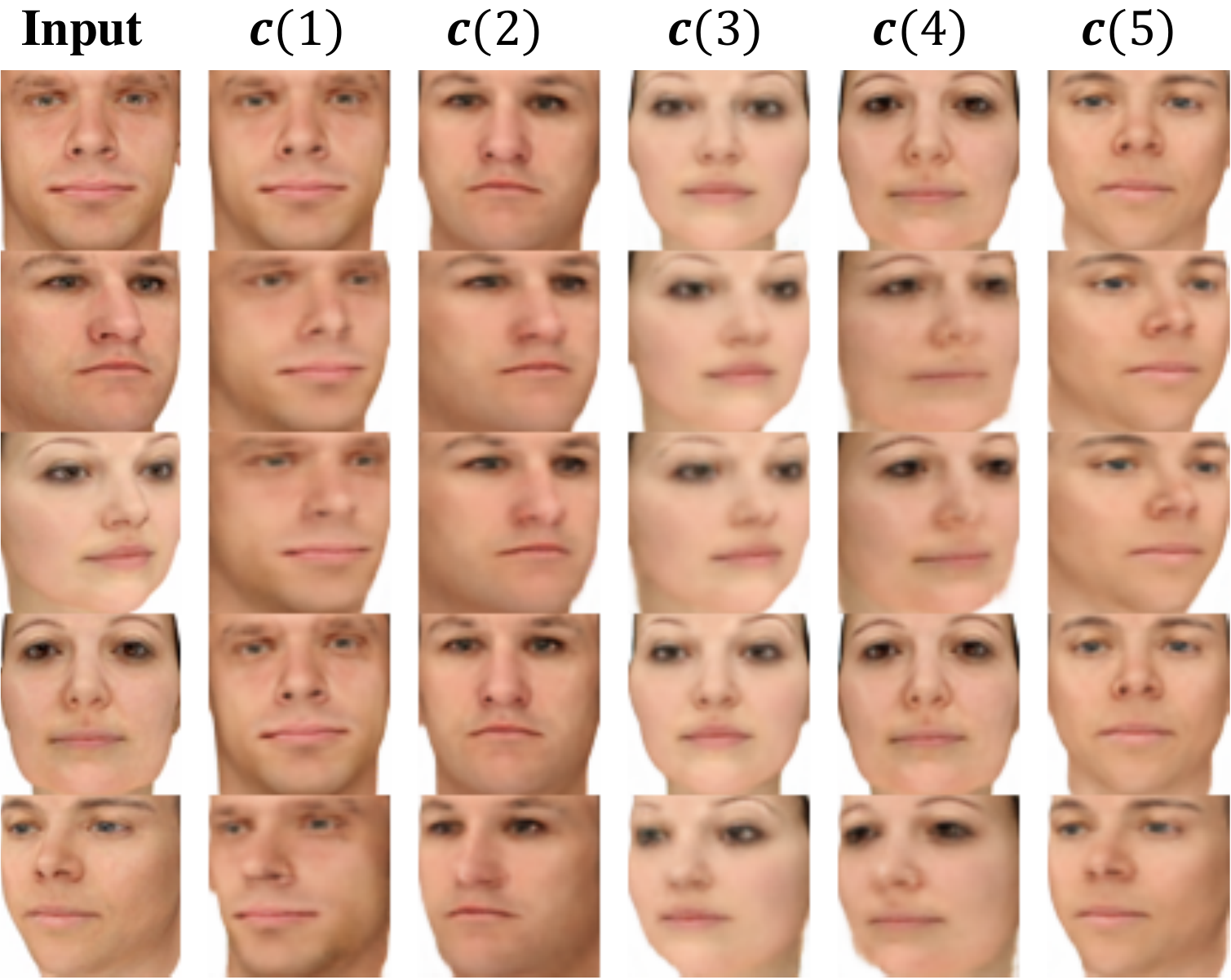}
		\subcaption{Image synthesis results for the proposed model}
		\label{fig:swap_sub1}
	\end{subfigure}
	\begin{subfigure}[b]{1\textwidth}
		\centering
		\includegraphics[width=.23\textwidth]{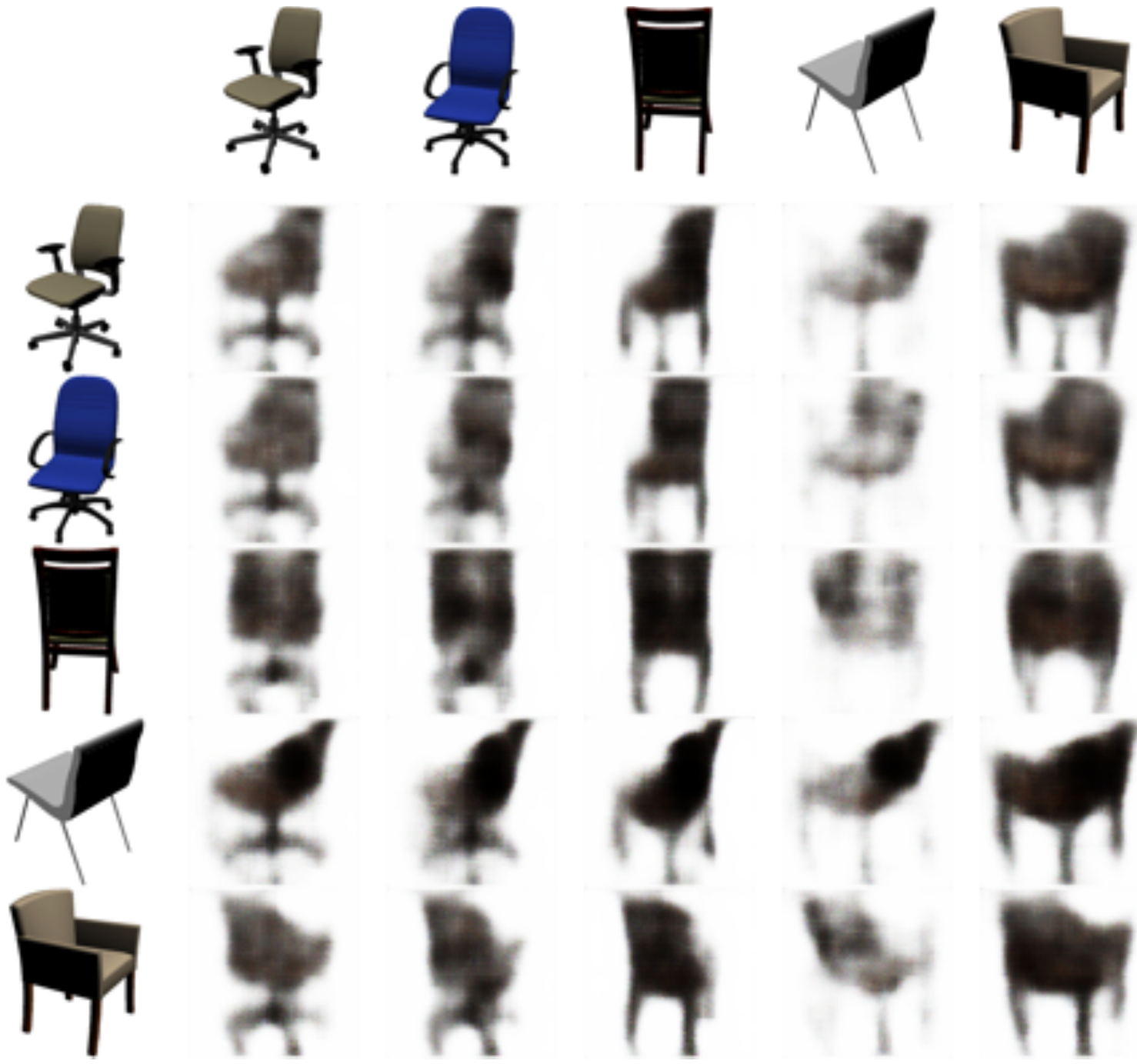}\qquad
		\includegraphics[width=.24\textwidth]{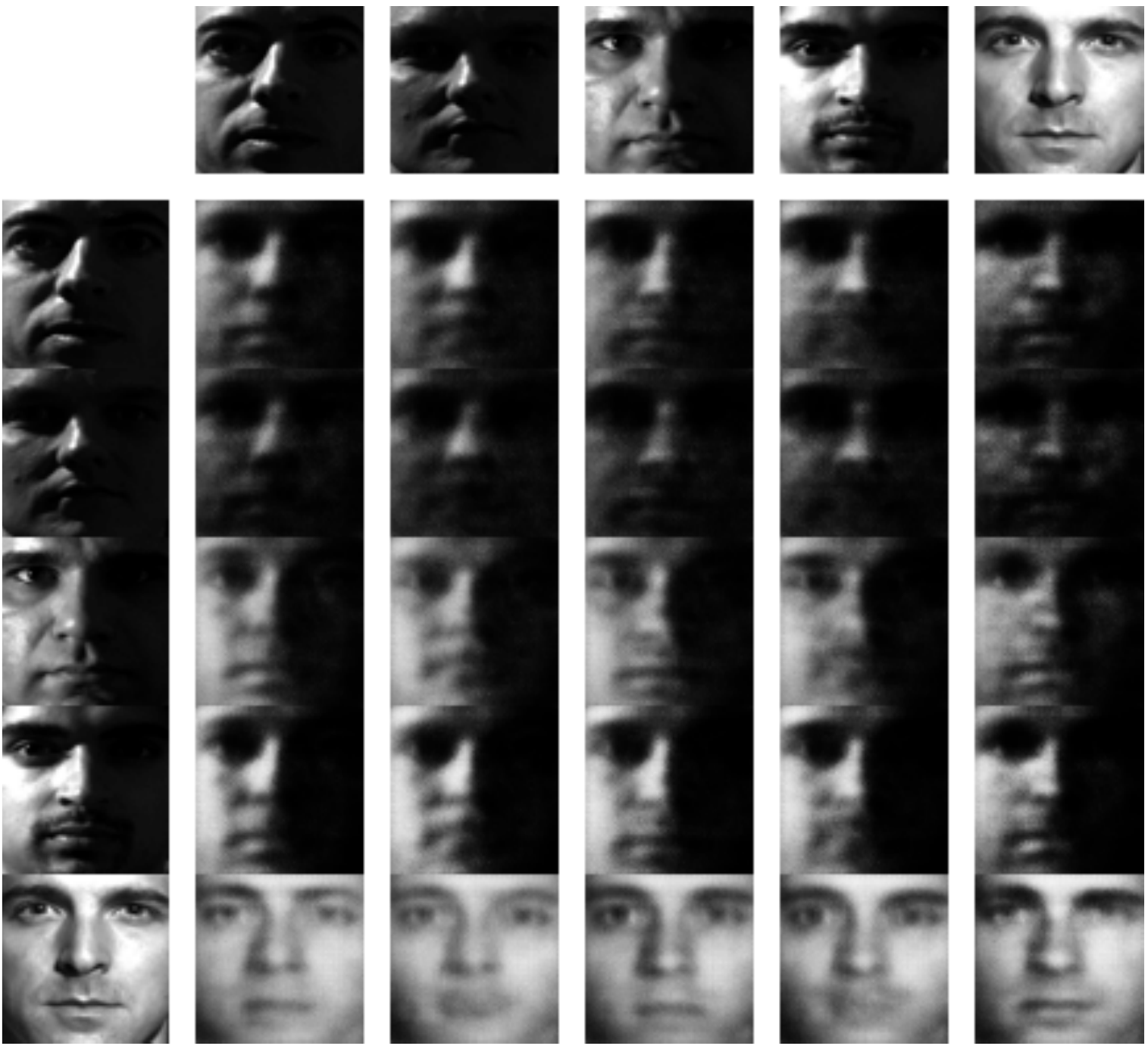}\qquad
		\includegraphics[width=.24\textwidth]{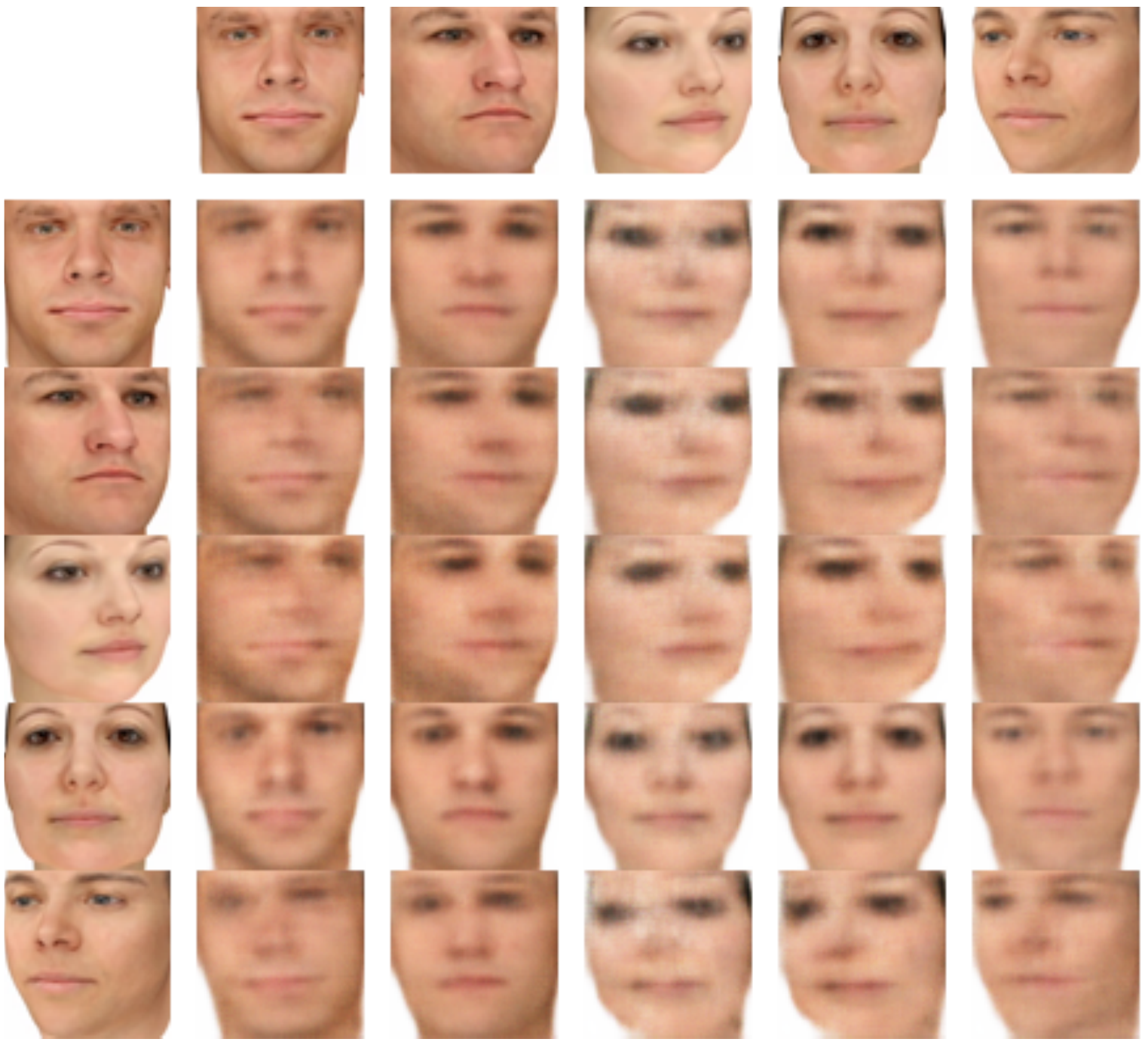}
		\subcaption{Image synthesis results for the model in~\cite{hadad2018two}}
		\label{fig:swap_sub2}
	\end{subfigure}
	\begin{subfigure}[b]{1\textwidth}
		\centering
		\includegraphics[width=.23\textwidth]{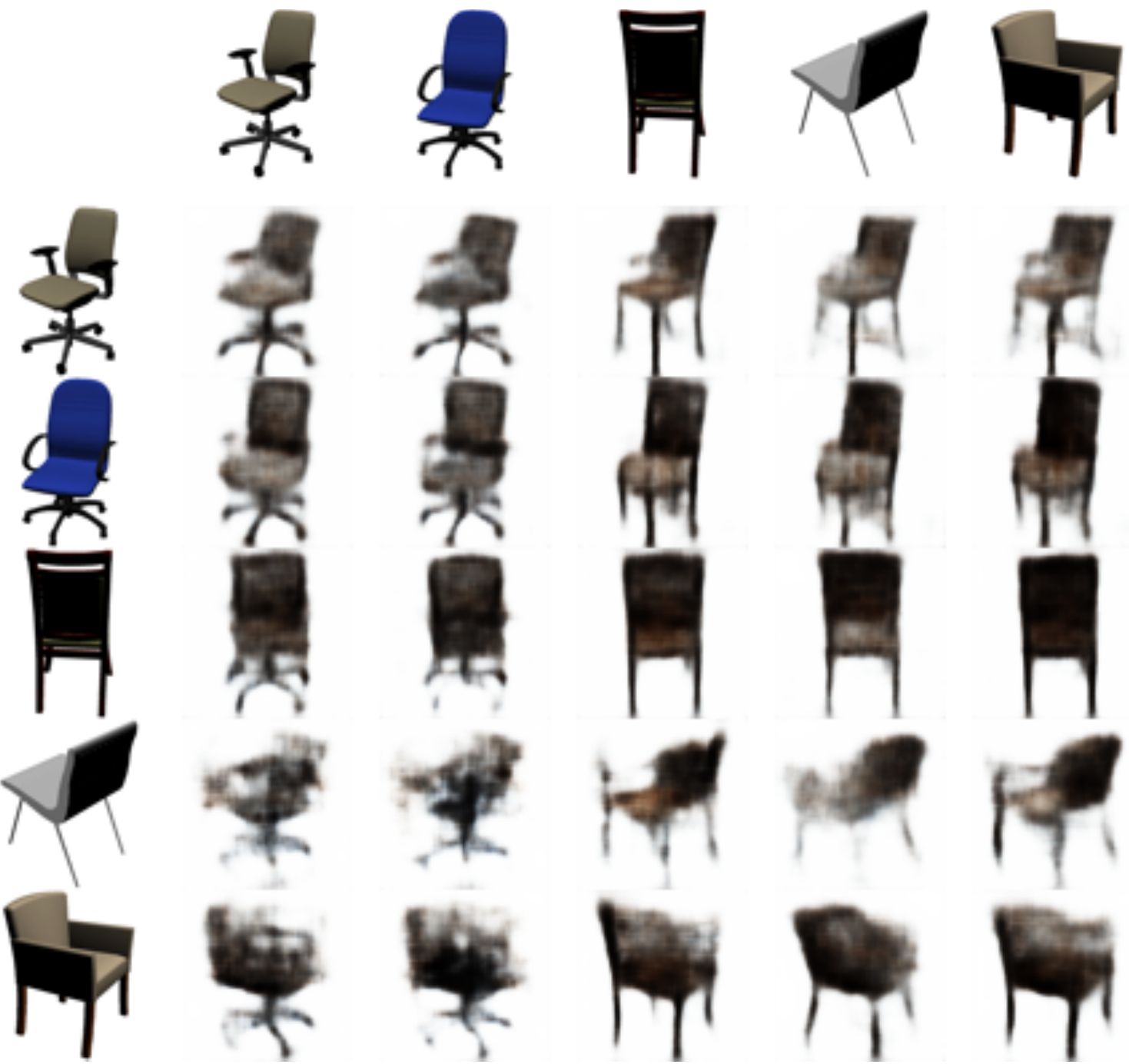}\qquad
		\includegraphics[width=.23\textwidth]{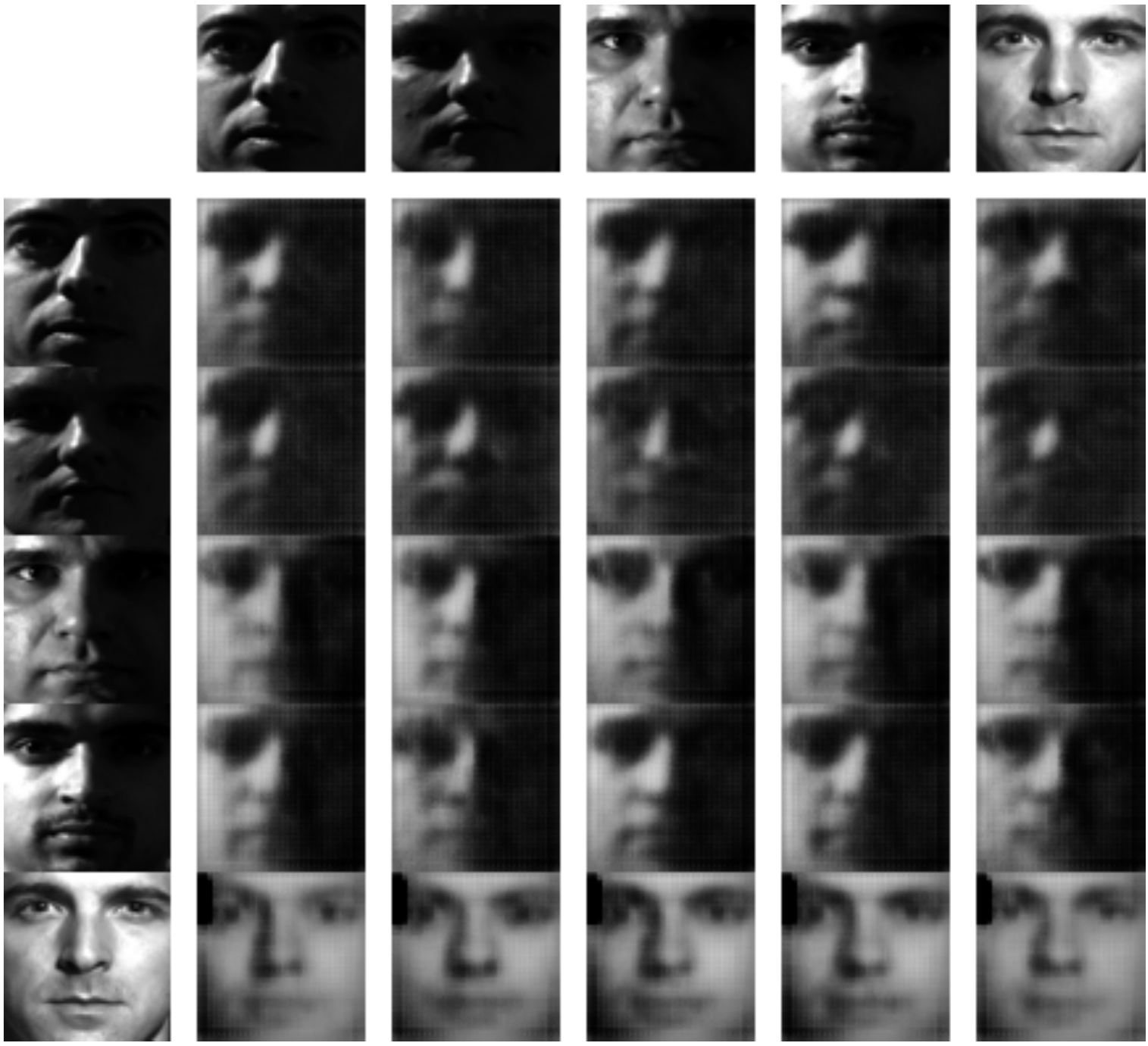}\qquad
		\includegraphics[width=.24\textwidth]{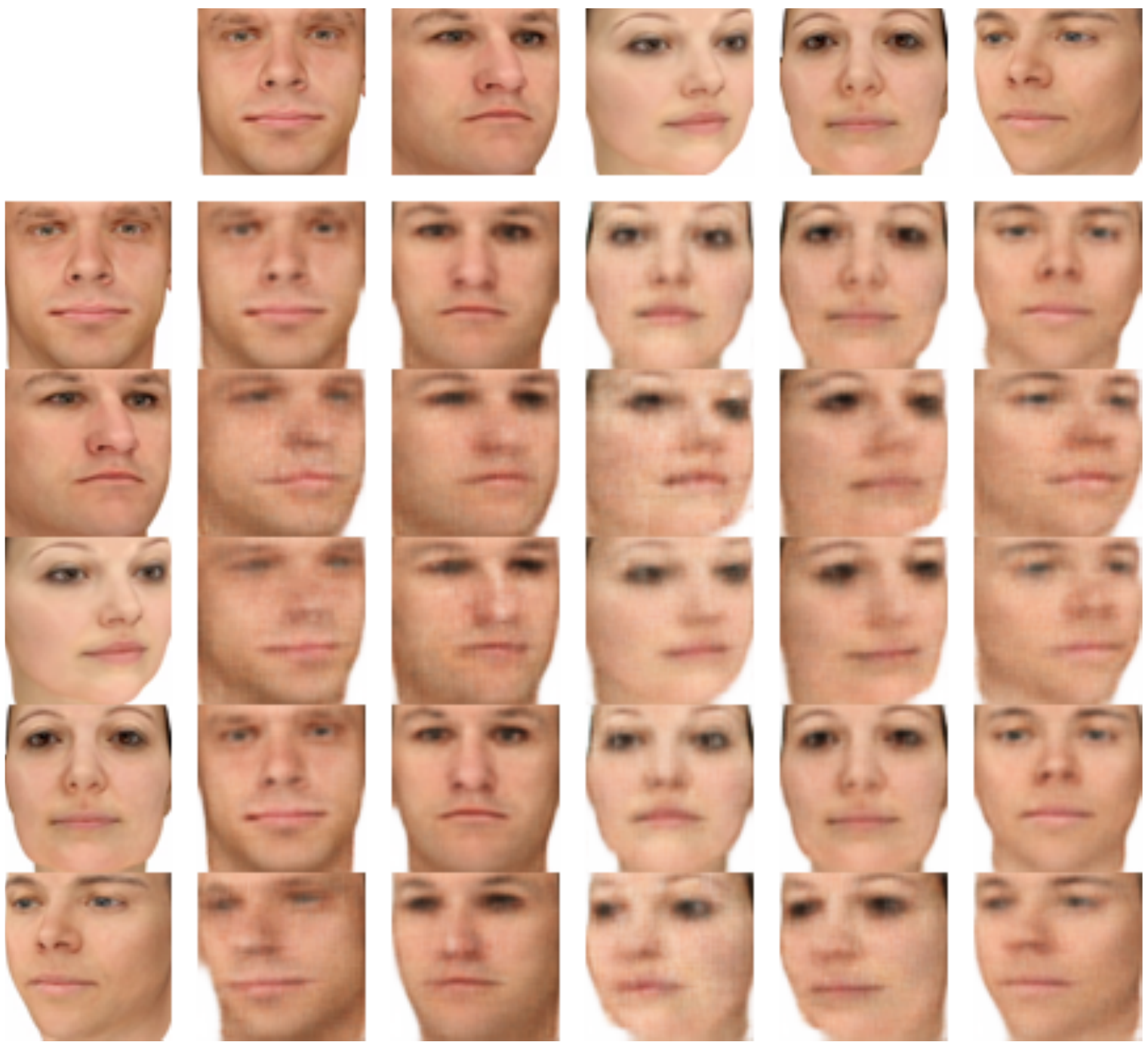}
		\subcaption{Image synthesis results for the model in~\cite{harsh2018disentangling}}
		\label{fig:swap_sub3}
	\end{subfigure}
	\caption{Image synthesis by altering the specified factor of variation for 3D Chairs, YaleFace and UPNA Synthetic datasets (from left to right). 
	}
	\vglue -0.70cm
	\label{fig:swap}
\end{figure*}

\begin{table}[!htb]
\caption{Comparison of \textit{GLS} values for the competing models. 
$\uparrow$ means higher is better. $\downarrow$ means lower is better.
}
\vglue -0.2cm
\label{tbl:GLS_table}
\centering
\resizebox{1\columnwidth}{!}{
	\setlength\extrarowheight{3pt}
	\begin{tabular}{c c c c c c}
		\toprule
		\multirow{2}{*}{\textbf{Datasets}} &  \multirow{2}{*}{\textbf{Factors of variation}} & \multicolumn{3}{ c }{\textbf{Methods}}  \\ 
		\cline{3-6}
		&   & ~\cite{hadad2018two}
		& ~\cite{harsh2018disentangling} & Ours & Ours w/o b.w. cycle  \\
		\cline{1-6}
		\multirow{3}{*}{3D Chairs} & Chair Style $\uparrow$ & 0.02 & 0.02  & \textbf{0.87} & 0.77 \\
		& $\theta$  $\uparrow$ & 0.56 & 0.61 & \textbf{0.66} & 0.66\\
		& $\phi$ $\uparrow$ & 0.38 & 0.49 & \textbf{0.57} & 0.52 \\
		\cline{1-6}
		\multirow{2}{*}{YaleFace} &Identity $\uparrow$& 0.24 & 0.07 & \textbf{0.98} & 0.97\\
		&  Illumination Cond. $\uparrow$ & 0.17 & 0.29 & \textbf{0.70} & 0.68\\
		\cline{1-6}
		\multirow{4}{*}{UPNA } & Identity $\uparrow$ &0. 88 & 0.98 & \textbf{1.00} & 0.99\\
		& Yaw $\downarrow$ & 3.51 & 2.65 & \textbf{2.55} & 2.62\\
		& Pitch $\downarrow$ & 4.07  & 2.84  & \textbf{2.46} & 2.95 \\
		& Roll $\downarrow$ & 3.17 & 1.47  & \textbf{1.37} & 1.39\\
		\bottomrule
	\end{tabular}}	
	\vglue -0.2cm
\end{table}
Table~\ref{tbl:GLS_table} reports the \textit{GLS} for the three datasets. 
We first observe that the proposed model consistently achieves better scores compared to the benchmark models. 
We can also see that the backward cycle does indeed improve the quality of the synthesized images.
In particular, \textit{GLS} values for the specified factors (chair style and identity) for our model are nearly perfect suggesting that our model manages to accurately alter the specified factor value in the generated images.   
With respect to unspecified factors of variation, our model yields a high \textit{GLS} value for the illumination condition ($0.70$) and a low value for head pose (e.g., $1.37$ for roll angle). 
While the achieved scores on viewing orientation ($\theta$, $\phi$) for our model are slightly lower than expected, they are still better than those for the benchmarks.
This is likely because our model occasionally fails to precisely construct chairlegs or arms (see Fig.~\ref{fig:swap_sub1}), which provide important cues for recognizing the viewing orientation.
It is worth mentioning that the performance differences are less significant on UPNA Synthetic dataset.
One possible reason is that it has the maximum number of training samples per class among the three datasets which could benefit the training of the generator.
%
%
%

As can be seen in Figure~\ref{fig:swap}, our model can change a specified factor of variation in an input image,  such as face identity or chair style, by adjusting class code $\mathbf{c}$.
Meanwhile, the other unspecified factors such as orientation, illumination condition or head pose of the input image are largely preserved in its synthetic version. 
Overall, images generated by our model are realistic although distortions may occur in image details, e.g.,  chair legs (see the fifth image in the second row of Fig.~\ref{fig:swap_sub1}). 
In contrast, the benchmark methods can only combine the specified factors from one source image and the unspecified factors from another source image to generate a new image. 
Therefore, they have less flexibility to modify a  specified factor of variation to a desired value.
Images shown in Figs.~\ref{fig:swap_sub2} and ~\ref{fig:swap_sub3} are generated by feeding the specified representations from images in the first row, and the unspecified representations from images in the first column to the decoder.
The visual quality of corresponding images is inferior to those from our model; blur and distortions are clearly visible. 
Furthermore, the benchmark methods are less effective in maintaining certain important factors of variation, e.g., color in the synthesized images (see the generated chair images in Figs.~\ref{fig:swap_sub2} and~\ref{fig:swap_sub3}). 

The remarkable consistency of the quantitative and qualitative results confirms the effectiveness of the proposed model in creating realistic images with a desired value for the specified factor and the same unspecified traits as the source images. 

\begin{figure}[!htb]
	\vglue -0.8cm
	\centering
	\begin{subfigure}[!htb]{0.24\textwidth}
		\centering
		\includegraphics[width=1\linewidth]{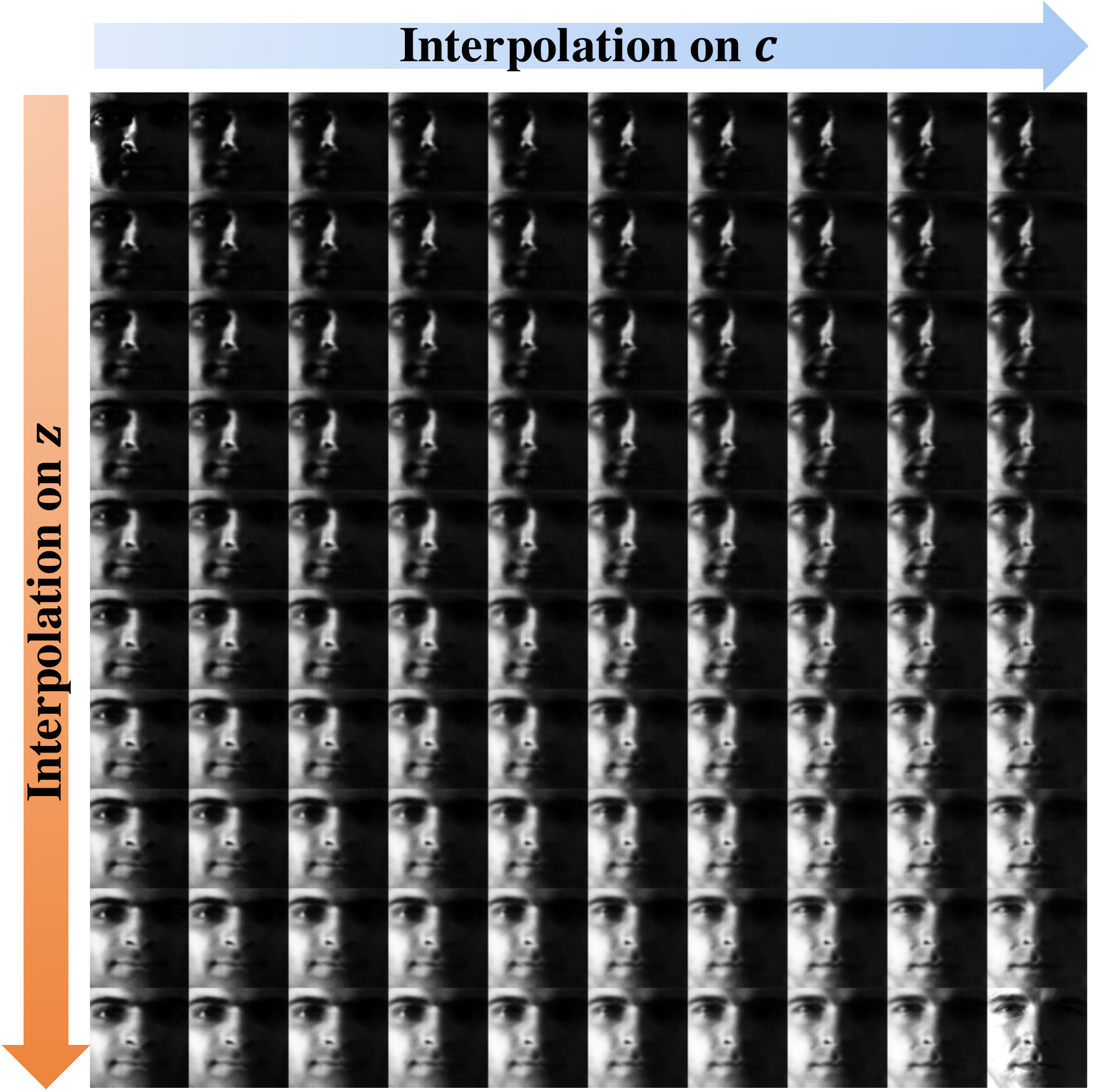}
		\caption{YaleFace}
		\label{fig:yaleface_morph} 
	\end{subfigure}%
	\begin{subfigure}[!htb]{0.24\textwidth}
		\centering
		\includegraphics[width=1\linewidth]{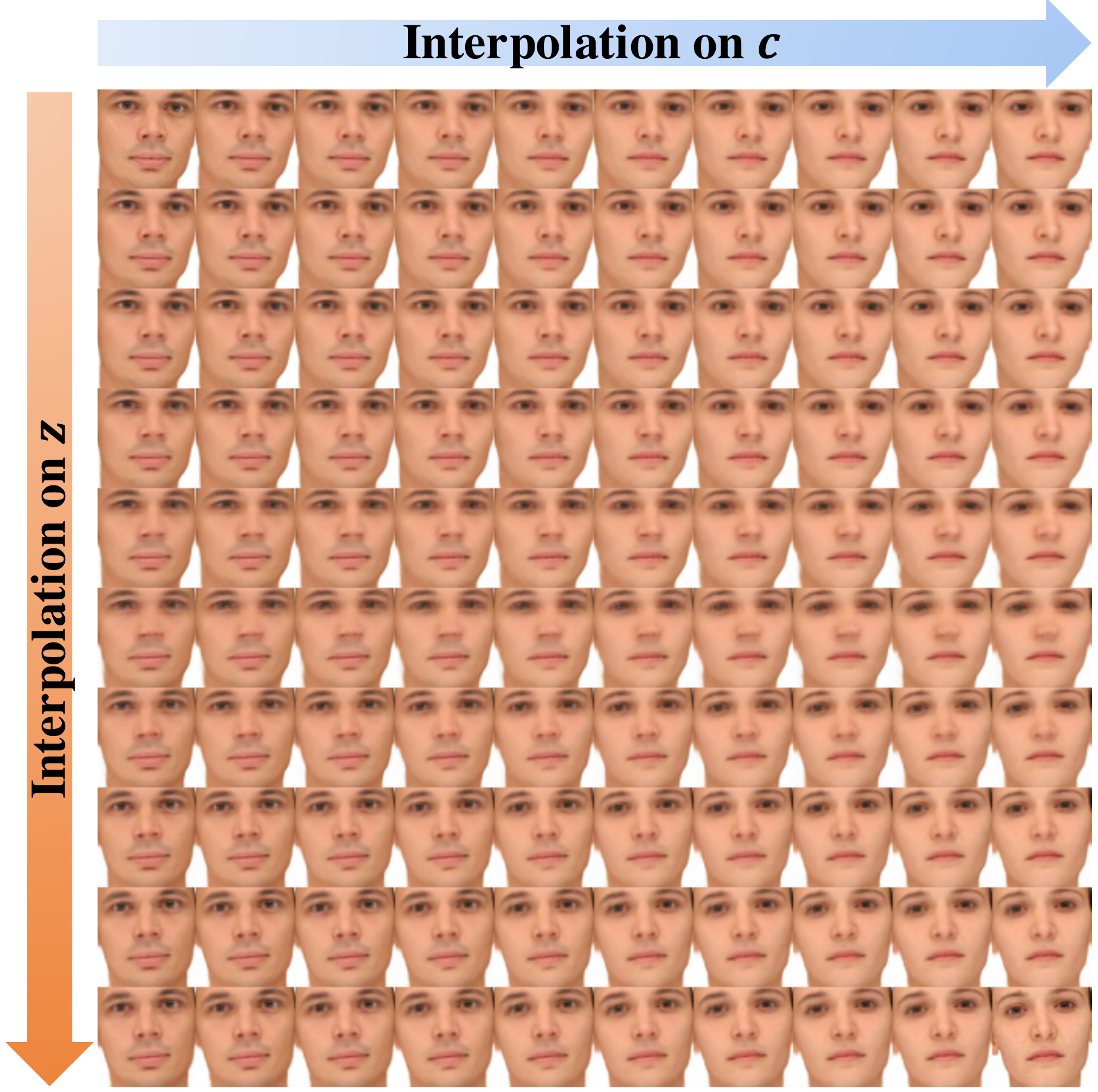}
		\caption{UPNA Synthetic}
		\label{fig:upna_morph}
	\end{subfigure}
	\vglue -0.3cm
	\caption{Linear interpolation results for the proposed model in the latent space ($\mathbf{z}$) and class code space ($\mathbf{c}$). The top-left and bottom-right images are taken from the test set.  }
	\label{fig:dense_morph}
	\vglue -2.0cm
\end{figure}

\noindent \textbf{Interpolation of synthesis variables:}
In order to further evaluate the generative capacity of the proposed model, we conducted additional experiments wherein we linearly interpolate between latent representations and class codes of an initial and a final image in order to obtain 
a series of new image representations and class codes which are then combined and fed to a trained decoder to synthesize new images. 
Specifically, let $\mathbf{z}_{\mathrm{initial}}, \mathbf{z}_{\mathrm{final}}$ and 
$\mathbf{c}_{\mathrm{initial}}, \mathbf{c}_{\mathrm{final}}$ denote, respectively, the learned latent representations and class codes of the initial and a final images and
$
\mathbf{c}_{\mathrm{interp}} = (1 - \alpha_c)\mathbf{c}_{\mathrm{initial}} + \alpha_c \mathbf{c}_{\mathrm{final}}
$
and
$
\mathbf{z}_{\mathrm{interp}} = (1 - \alpha_z)\mathbf{z}_{\mathrm{initial}} + \alpha_z \mathbf{z}_{\mathrm{final}}
$
their interpolated values, where $\alpha_c, \alpha_z \in[0,1]$. We synthesize new images by passing $(\mathbf{c}_{\mathrm{interp}}, \mathbf{z}_{\mathrm{interp}})$ into the decoder.
Surprisingly, when this is applied to a face dataset, our trained model can generate a sequence of face images that show a seamless transition from one identity into another, i.e., face morphing (\ref{fig:dense_morph}), and also a seamless transition from one value of an unspecified factor (e.g., illumination, pose) into another (columns of Fig.~\ref{fig:dense_morph}). This is despite the fact that the model can only see one-hot codes specifying \textit{discrete} identities during training.
%
%
In Fig.~\ref{fig:dense_morph}, the class code is constant within each column while the representation is constant within each rows.
We observe that when interpolating $\mathbf{c}$, the unspecified factors such as illumination or head pose are consistent, while the specified factor (identity) changes gradually.
In contrast, when interpolating $\mathbf{z}$ the specified factor remains unchanged but the unspecified factors transform continuously.

\section{Conclusion}
This paper presents a conditional adversarial network for learning an image representation that is invariant to a specified factor of variation, while 
maintaining unspecified factors. 
The proposed model does not produce degenerate solutions due to a novel cyclic forward-backward training strategy.
%
Quantitative results from a broad set of experiments show that our model performs better or equally well compared to two state-of-the-art methods in
learning invariant image representations.
Once trained, our model is also generative as it enables synthesis of a realistic image having a desired value for the specified factor.
Both qualitative and quantitative evaluation results confirm that our model can produce better quality images than the competing models. 

\section{Acknowledgement}
This work was supported by the NSF under Lighting Enabled Systems and Applications ERC Cooperative Agreement No.EEC-08120256

{\small
\bibliographystyle{ieee_fullname}
\bibliography{egbib}
\pagebreak

\appendix
\section{Supplementary Material}
In this supplementary material, section~\ref{sec:ablation} provides an ablation study which evaluates the impacts of various training objectives in our framework.
Section~\ref{sec:image_syn} gives examples of synthesized images from our trained model without using an input image. 

\subsection{Ablation study}
\label{sec:ablation}
\begin{table*}[!h]
\caption{Comparison of the quality of invariance of representations generated by the proposed model trained with various objectives. 
$\uparrow$ means higher is better. $\downarrow$ means lower is better.}
\vglue -0.1cm
\label{tbl:ablation_1}
\centering
\resizebox{1.0\textwidth}{!}{
\setlength\extrarowheight{3pt}
\begin{tabular}{c c c c c c c }
	\toprule
	\multirow{2}{*}{\textbf{Dataset}} &  \multirow{2}{*}{\textbf{Factors of variation}} & \multicolumn{5}{ c }{\textbf{Methods}}  \\ 
	\cline{3-7}
	&   &  Random  guess/ & \multirow{2}{*}{Forward cycle  }
	& \multirow{2}{*}{Forward cycle + } & \multirow{2}{*}{Forward cycle +  }
	& \multirow{2}{*}{Forward cycle + } \\
	& & Median  &  & cycle-consistency on \textbf{z} & cycle-consistency on \textbf{x} &  cycle-consistency on \textbf{x} \& \textbf{z}\\
	\cline{1-7}
	\multirow{3}{*}{3D Chairs} & Chair Style $\downarrow$ & $0.07\%$ & $3.21\%$ & 0.30$\%$ & 3.51$\%$ & 0.79$\%$ \\
	& $\theta$  $\uparrow$ & 50$\%$ & 74.37$\%$& 72.00$\%$ & 84.40$\%$  & 78.17$\%$ \\
	& $\phi$ $\uparrow$ & 3.22$\%$ & 69.54$\%$ & 70.96$\%$ & 72.60$\%$ & 71.90$\%$ \\
	\cline{1-7}
	\multirow{2}{*}{YaleFace} &Identity $\downarrow$& 2.63$\%$ & 12.36$\%$ & 4.47$\%$ & 17.89$\%$ & 6.97$\%$ \\
	&  Illumination Cond. $\uparrow$ & 1.56$\%$ & 85.40$\%$ & 82.37$\%$ & 85.73$\%$ & 85.50$\%$ \\
	\cline{1-7}
	\multirow{4}{*}{UPNA Synthetic} & Identity $\downarrow$ & 10$\%$ & 33.40$\%$ & 15.61$\%$ & 56.48$\%$ & 18.05$\%$ \\
	& Yaw $\downarrow$ & 5.10$^{\circ}\pm$6.70$^{\circ}$ & 2.10$^{\circ}\pm$2.08$^{\circ}$ & 2.48$^{\circ}\pm$2.60$^{\circ}$  & 1.65$^{\circ}\pm$1.57$^{\circ}$ & 2.12$^{\circ}\pm$2.12$^{\circ}$  \\
	& Pitch $\downarrow$ & 4.98$^{\circ}\pm$5.02$^{\circ}$  & 2.20$^{\circ}\pm$2.06$^{\circ}$  & 2.82$^{\circ}\pm$3.12$^{\circ}$  & 1.73$^{\circ}\pm$1.62$^{\circ}$ & 2.23$^{\circ}\pm$2.10$^{\circ}$ \\
	& Roll $\downarrow$ & 4.68$^{\circ}\pm$6.88$^{\circ}$ & 1.29$^{\circ}\pm$1.43$^{\circ}$  & 1.69$^{\circ}\pm$1.99$^{\circ}$  & 0.86$^{\circ}\pm$0.85$^{\circ}$ & 1.16$^{\circ}\pm$1.24$^{\circ}$ \\
	\bottomrule
\end{tabular}}	
\end{table*}
\bigskip

\begin{table*}[!h]
\caption{Comparison of \textit{GLS} values for the proposed model trained with various objectives. 
$\uparrow$ means higher is better. $\downarrow$ means lower is better.
}
\vglue -0.2cm
\label{tbl:ablation_2}
\centering
\resizebox{1.8\columnwidth}{!}{
	\setlength\extrarowheight{3pt}
	\begin{tabular}{c c c c c c}
		\toprule
		\multirow{2}{*}{\textbf{Datasets}} &  \multirow{2}{*}{\textbf{Factors of variation}} & \multicolumn{3}{ c }{\textbf{Methods}}  \\ 
		\cline{3-6}
	&   &  \multirow{2}{*}{Forward cycle  }
	& \multirow{2}{*}{Forward cycle + } & \multirow{2}{*}{Forward cycle +  }
	& \multirow{2}{*}{Forward cycle + } \\
	&  &  & cycle-consistency on \textbf{z} & cycle-consistency on \textbf{x} & cycle-consistency on \textbf{x} \& \textbf{z}\\
		\cline{1-6}
		\multirow{3}{*}{3D Chairs} & Chair Style $\uparrow$ & 0.77 &  0.76 & 0.92 & 0.87 \\
		& $\theta$  $\uparrow$ & 0.66 & 0.64 & 0.78 & 0.66 \\
		& $\phi$ $\uparrow$ & 0.52 & 0.50 & 0.59 & 0.57  \\
		\cline{1-6}
		\multirow{2}{*}{YaleFace} &Identity $\uparrow$& 0.97 & 0.98 & 0.98 & 0.98 \\
		&  Illumination Cond. $\uparrow$ & 0.68 & 0.59 & 0.72 & 0.70\\
		\cline{1-6}
		\multirow{4}{*}{UPNA Synthetic} & Identity $\uparrow$ &0.99 & 0.99 & 1.00 & 1.00 \\
		& Yaw $\downarrow$ & 2.62 & 3.13 & 2.40 & 2.55 \\
		& Pitch $\downarrow$ & 2.95  & 3.58  & 2.30 & 2.46   \\
		& Roll $\downarrow$ & 1.39 & 2.14  & 1.28 & 1.37 \\
		\bottomrule
	\end{tabular}}	
\end{table*}

In this section, we perform an ablation study for various training objectives for the proposed model on three datasets (see corresponding results in Tables~\ref{tbl:ablation_1} and~\ref{tbl:ablation_2}). 
Specifically, we examine the following cases:
\begin{enumerate}
  \item training the model with forward cycle only (using Eq.(1) and Eq.(2) in the paper),
  \item training the model with forward cycle and cycle-consistency constraint on the latent space (using Eq.(1), Eq.(2) and the first term in Eq.(3) in the paper),
  \item training the model with forward cycle and cycle-consistency constraint on the image space (using Eq.(1), Eq.(2) and the second term in Eq.(3) in the paper),
  \item training the complete model with both forward and backward cycles.
\end{enumerate}

We first observe that the cycle-consistency constraint on the latent space alone significantly improves invariance, e.g., identity classification CCR decreases from 33.40\% (forward cycle only) to 15.61\% for UPNA synthetic dataset.
However, it also slightly diminishes information about unspecified factors of variation, which is reflected in both latent and image spaces, e.g., 
head pose estimation errors computed from the latent representations and GLS values of the pose angles are $0.47\degree$ and 0.63 larger than those of case 1 (in average), respectively.
%
This is likely because the specified and certain unspecified factors of variation are not strictly independent, e.g., some subjects have a wider range of pose angles in UPNA dataset. 
Thus, when the model is trained to disentangle the specified factor of variation from the latent space, the correlated unspecified factors are inevitably affected. 

When combining a forward cycle with cycle-consistency constraint on the image space, 
we observe that the learned representations and synthesized images retain more information about the original image compared to other cases, e.g., the model in this case yields the best classification CCRs and GLS scores for the viewing orientations for 3D chairs. 
%
However, it also yields the highest classification CCR for chair style, a specified factor, which
%
indicates that although the model is provided with an input class code $\mathbf{c}$ to control the specified factor of variation in the synthesized image, the image reconstruction objective still leads the model to preserve a certain amount of the original information about the  specified factor in the latent representation.

In the case of training the model with a forward cycle and complete backward cycle, we witness that the model achieves a good balance between eliminating information about the specified factor and preserving information about the unspecified factors of variation in the latent space.
More importantly, the model in this case consistently outperforms the model merely trained with a forward cycle in terms of the ability to suppress classification performance for the specified factor, while ensuring that the estimation performance for the unspecified factors does not fall short. 
Similarly, the synthesized images from the complete model consistently have a better GLS scores than the images from the model trained with forward cycle only. 
%
Such observations justify the effectiveness and indispensability of both cycle-consistency constraints.

\subsection{ Image synthesis without input image}
\label{sec:image_syn}
Once trained, our model can also synthesize novel images without using an input image due to the constraint imposed over the latent space. 
To generate a new image, we first sample a latent vector from a prior distribution (in our experiments: $\mathbf{z} \sim \mathcal{N}(\mathbf{0}, \mathbf{I})$). 
Then we concatenate it with a class code and feed them into a trained decoder to synthesize a new image. 
As shown in Fig.~\ref{fig:img_from_z}, the synthesized images are realistic. 
Although our model cannot control the unspecified factors without an input image, the synthesized images from randomly sampled latent vectors are still useful for some applications, e.g., dataset augmentation. 
\begin{figure}[!h]
	\centering
	\includegraphics[width=1.0\linewidth]{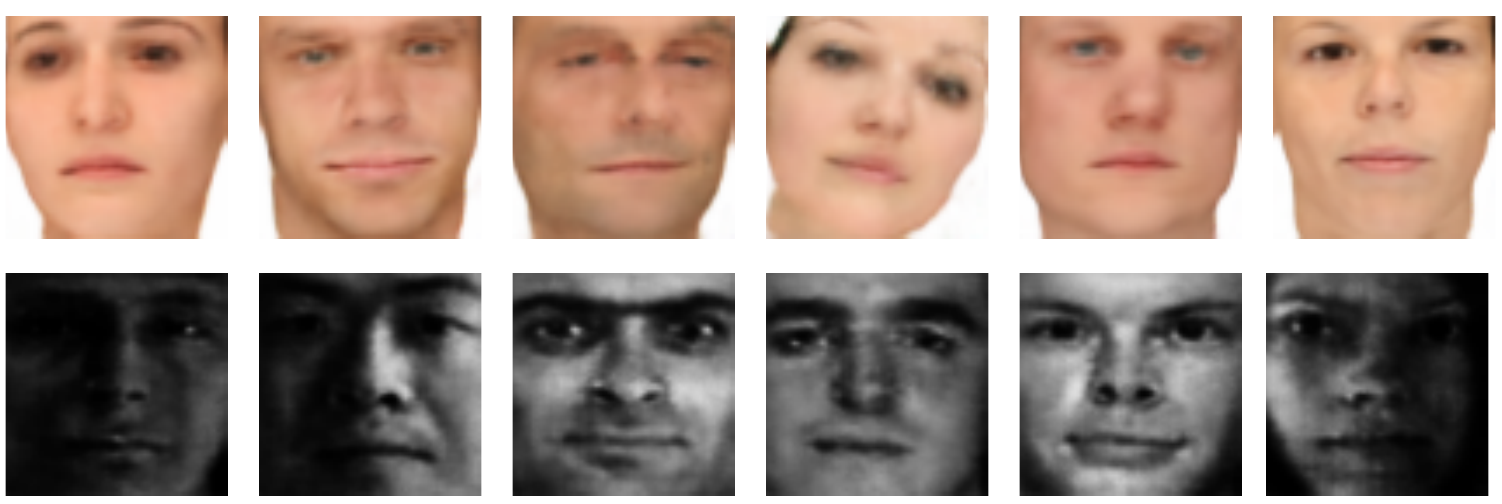}
	\caption{Image synthesis without input image; $\mathbf{z}$ is sampled from $\mathcal{N} (\mathbf{0}, \mathbf{I})$.}
	\label{fig:img_from_z}
\end{figure}

\end{document}